\documentclass[conference]{IEEEtran}
\IEEEoverridecommandlockouts
\usepackage{booktabs}
\usepackage{multirow}
\usepackage{cite}
\usepackage{amsmath,amssymb,amsfonts}
\usepackage{algorithmic}
\usepackage{graphicx}
\usepackage{textcomp}
\usepackage{xcolor}
\usepackage{algorithmic}
\usepackage{hyperref}
\usepackage{pifont}
\def\BibTeX{{\rm B\kern-.05em{\sc i\kern-.025em b}\kern-.08em
    T\kern-.1667em\lower.7ex\hbox{E}\kern-.125emX}}
\begin{document}

\title{Bridging the Perception-Cognition Gap:Re-engineering SAM2 with Hilbert-Mamba for Robust VLM-based Medical Diagnosis\\
}
\author{
\IEEEauthorblockN{Hao Wu$^{1\#}$,  Hui Li$^{2\#}$\thanks{\textsuperscript{\#}These authors contributed equally to this work.}, Yiyun Su$^{3}$}\IEEEauthorblockA{\IEEEauthorrefmark{1}Southern University of Science and Technology, Shenzhen, China}\IEEEauthorblockA{\IEEEauthorrefmark{2}School of Informatics, Xiamen University, Xiamen, China} \IEEEauthorblockA{\IEEEauthorrefmark{3}Rutgers University, New Brunswick, United States}}

\maketitle

\begin{abstract}
Abstract—Recent studies suggest that Visual Language Models (VLMs) hold great potential for tasks such as automated medical diagnosis. However, processing complex three-dimensional (3D) multimodal medical images poses significant challenges—specifically, the effective integration of complementary information and the occasional oversight of subtle yet critical pathological features. To address these issues, we present a novel two-stage fusion framework termed \textbf{Hilbert-VLM}. This framework leverages the \textbf{HilbertMed-SAM} module for precise lesion segmentation, with the generated multimodal enhanced prompts then guiding the VLM toward accurate disease classification. Our key innovation lies in the systematic redesign of the Segment Anything Model 2 (SAM2) architecture: we incorporate Hilbert space-filling curves into the scanning mechanism of the Mamba State Space Model (SSM) to maximize the preservation of spatial locality in 3D data, a property critical for medical image analysis. We also introduce a novel Hilbert-Mamba Cross-Attention (HMCA) mechanism and a scale-aware decoder to capture fine-grained details. Meanwhile, the prompt enhancement module unifies segmentation masks and their corresponding textual attributes into an information-dense prompt to support VLM inference. Extensive experiments were conducted to validate the effectiveness of the Hilbert-VLM model. On the BraTS2021 segmentation benchmark, it achieves a \textbf{Dice score of 82.35\%}, with a diagnostic classification accuracy (\textbf{ACC}) of \textbf{78.85\%}. These results demonstrate that the proposed model offers substantial potential to improve the accuracy and reliability of medical VLM-based analysis.
\end{abstract}

\begin{IEEEkeywords}
Vision-Language Models, Medical Image Segmentation, Multi-Modal Fusion, Hilbert Curve, Disease Classification
\end{IEEEkeywords}

\section{Introduction}

Accurate diagnosis of brain pathologies is critical for clinical evaluation and treatment\cite{kruser2019nrg}. Radiologists rely on multimodal MRI sequences (e.g., T1 and FLAIR) to synthesise complementary information\cite{liu2023deep}. However, this manual process is time-consuming and prone to inter-observer variability\cite{joskowicz2019inter,jungo2018effect}. Large visual-language models (VLMs), such as Qwen-VL, are poised to bring about a paradigm shift in automated diagnosis, as demonstrated by the work of Achiam et al.\cite{achiam2023gpt} and Bai\cite{bai2023qwen}, thanks to their advanced cross-modal understanding capabilities.

However, applying general-purpose VLMs, which are designed for 2D natural images, to 3D multimodal medical imaging remains challenging. For instance, these models cannot easily fuse the complex and heterogeneous spatial data from 3D scans. Secondly, pathological lesions tend to be small and the differences between them and healthy tissue are often very subtle. This may cause VLMs to overlook critical details\cite{dekoninck2023controlled}. Such perceptual failures, stemming from inadequate visual information fusion, limit the reliability of VLMs in high-stakes clinical applications\cite{awadalla2023openflamingo}.

Current approaches, such as fine-tuning or prompt engineering, do not address this fusion problem in a fundamental way. General-purpose segmentation models like SAM2\cite{ravi2024sam} also show limited performance without deep domain adaptation\cite{ma2024segment,zhang2024segment}. We posit that accurate diagnosis requires a framework that can first 'perceive accurately' and then 'reason correctly.' We propose Hilbert-VLM, a "segmentation-first, fusion-prompt" two-stage framework. First, we develop HilbertMed-SAM, a segmentation model designed for 3D multimodal imaging to achieve precise lesion delineation. Second, an innovative prompt module integrates the visual segmentation mask and textualized lesion attributes from HilbertMed-SAM into a unified, enhanced prompt to guide a VLM (we use Qwen-VL) for disease classification.

Our main contributions are: (i) A novel two-stage framework that integrates a Vision Foundation Model (VFM) and a VLM by decoupling `perception` and `cognition` tasks; (ii) The Hilbert-Mamba-Empowered SAM2 architecture, which re-engineers the encoder, memory, and decoder of SAM2 using Hilbert curves and Mamba SSMs for superior multimodal fusion and minute lesion detection; (iii) A Visual-Text Fusion Prompt mechanism that encodes and fuses the segmentation mask and lesion metadata to significantly improve VLM diagnostic accuracy; and (iv) SOTA performance on both lesion segmentation and disease classification tasks, offering a robust solution for AI-driven clinical decision support systems.
\iffalse

\fi
\section{Related Work}
\subsection{Foundation Models for Medical Image Segmentation}
Medical image segmentation is fundamental to clinical tasks. Architectures have progressed from CNNs (like 3D U-Net) to Transformer-based frameworks (like UNETR). More recently, the Segment Anything Model (SAM) introduced promptable segmentation, which led to medical-specific adaptations such as MedSAM. SAM2 incorporates additional memory-enhancing mechanisms, thereby unlocking new potential for 3D volume processing. However, we believe that the unique spatio-temporal nature of three-dimensional, multimodal medical data means that SAM2 requires more than just minor adjustments, it needs a fundamental architectural overhaul.
\subsection{Vision-Language Models in Clinical Applications}
Early visual language models (VLMs), such as CLIP, rely on large-scale paired image-text datasets. However, these datasets are difficult to obtain in medical contexts due to strict privacy restrictions and high costs. This has driven adaptations of general-purpose VLMs (e.g., Qwen-VL, Med-PaLM) using limited in-domain data. However, these adapted models often fail to recognise subtle details, which can result in misdiagnoses. To address this, we require two advances: an upstream model for precise visual perception, and a mechanism to convert visual discoveries, such as our segmentation masks, into a format suitable for VLM-based reasoning. Our multimodal fusion prompt module is designed to accomplish this task.
\subsection{State-Space Models and Locality Preservation in Vision}
State Space Models (SSMs), such as Mamba, are considered to be an effective alternative to transformers. Mamba is inherently one-dimensional. When combined with standard image flattening, it disrupts the spatial locality of 3D data, which hinders its application in medical imaging. Hilbert space-filling curves solve this by mapping 3D data to 1D while preserving spatial proximity. This allows Mamba to retain spatiotemporal information. We integrated this Hilbert-guided scanning with Mamba to redesign SAM2's encoder, memory, and decoder. The resulting architecture balances computational efficiency with high-fidelity spatial awareness, overcoming the earlier perception failures.

\begin{figure}[!t]
\centering
\includegraphics[width=0.9\columnwidth]{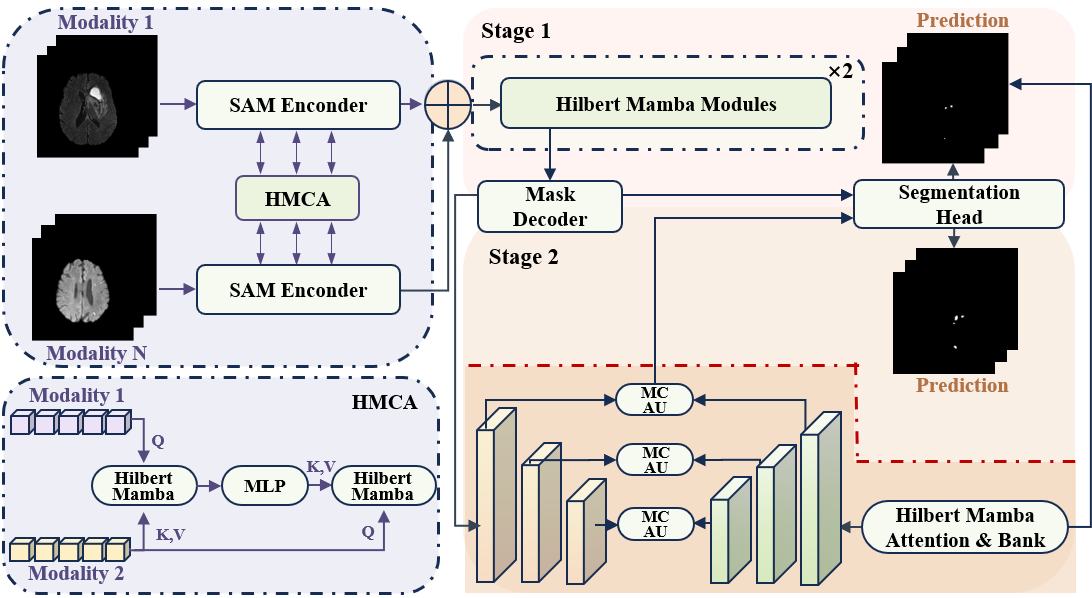}
\caption{The architecture of HilbertMed-SAM encompasses multimodal input and HMCA fusion, the Hilbert Mamba optimisation module and a dual-path (Stage 1/Stage 2) decoder for the final prediction.}
\label{fig_framework}
\end{figure}

\section{Methodology}

\subsection{Overall Framework of Hilbert-VLM}
Hilbert-VLM is a "segment-then-prompt" two-stage framework (Figure~\ref{fig_framework}). Stage 1 involves precise lesion segmentation. Stage 2 then uses these results to guide VLM for disease classification.

\subsection{HilbertMed-SAM: Re-engineering SAM2 for Medical Imaging}
The cornerstone of this framework is HilbertMed-SAM, which has been redesigned for multimodal 3D medical imaging (see Figure~\ref{fig_framework}). It comprises a foundational Hilbert-Mamba block, a multimodal encoder, memory-infused enhancement modules, and a dual-path decoder.

\subsubsection{Foundational Principle: The Hilbert-Mamba Block}
\begin{figure}[!t]
\centering
\includegraphics[width=0.9\columnwidth]{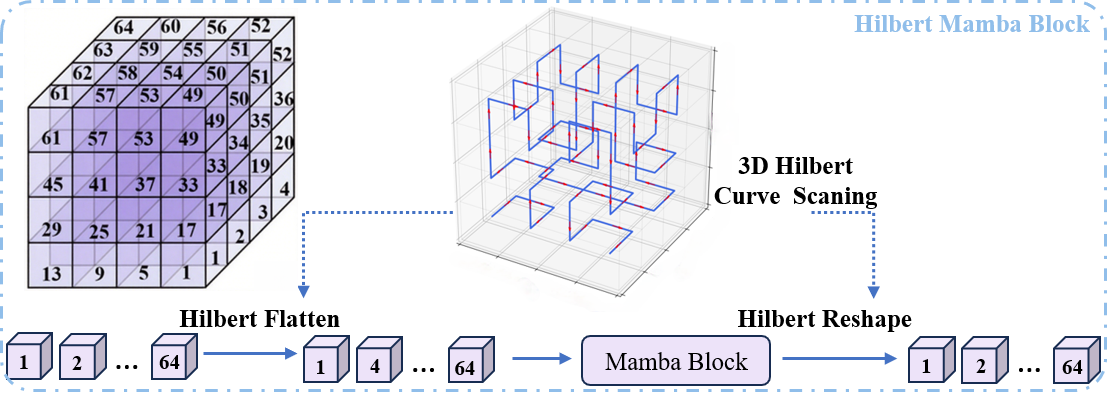}
\caption{The Hilbert Mamba block is based on a particular scanning principle. Unlike standard grating scans, which destroy spatial locality, our approach uses a 3D Hilbert curve scanning strategy.}
\label{fig_HMB}
\end{figure}

The Hilbert-Mamba Block (Figure~\ref{fig_HMB}) prevents locality issues in standard serialisation. A Hilbert Flatten uses a Hilbert curve to map 3D features to 1D, preserving relationships. This locality-aware sequence is processed by a Mamba block, whose State-Space Model (SSM) core efficiently models long-range dependencies while leveraging the preserved local context.

\subsubsection{Multimodal Encoder with HMCA}
The encoding process (Figure~\ref{fig_framework}) begins by passing each modality through a separate SAM Encoder. To merge complementary information, we introduce the HMCA module. First, the Key and Value ($K,V$) representations from the second modality are contextualized by a Hilbert-Mamba block and refined by an MLP. Second, the Query ($Q$) from the first modality interacts with this enriched context in another Hilbert-Mamba block to produce cross-fused features. This output is then combined with the first modality's features via element-wise addition.

\subsubsection{Memory-infused Hilbert-Mamba Modules}
\begin{figure}[!t]

\centering

\includegraphics[width=0.9\columnwidth]{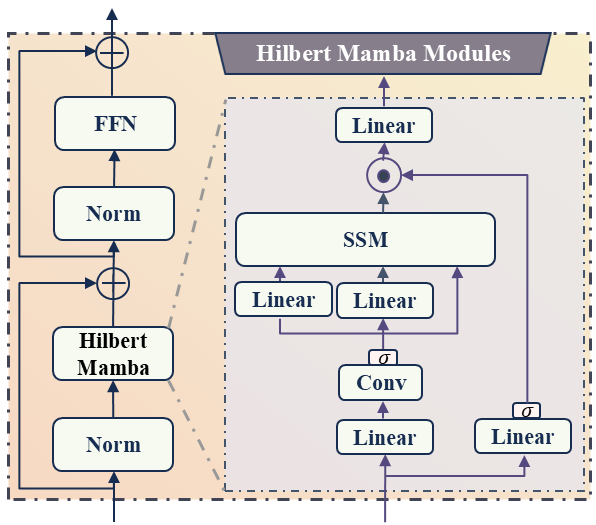}

\caption{Detailed architecture of the Hilbert Mamba Module.}

\label{fig_HMM}

\end{figure}

Fused features are processed slice-by-slice by $L=2$ Memory-infused Hilbert-Mamba Modules (Figure~\ref{fig_HMM}). These modules perform intra-slice spatial refinement (via HMB and FFN) and inter-slice spatiotemporal context propagation. They employ a GRU-like gating mechanism to update a memory state $M_t$ based on the previous state $M_{t-1}$ and current slice feature $f^t$:
\begin{align}
    u_t &= \sigma(W_u [f^t, M_{t-1}]) \\
    r_t &= \sigma(W_r [f^t, M_{t-1}]) \\
    \tilde{M}_t &= \tanh(W_m [f^t, r_t \odot M_{t-1}]) \\
    M_t &= (1 - u_t) \odot M_{t-1} + u_t \odot \tilde{M}_t
\end{align}
This mechanism selectively passes relevant context across slices to the decoder.

\subsubsection{Dual-Path Decoder}
The dual-path decoder (Figure~\ref{fig_framework}) is designed for high-precision reconstruction. We systematically replace standard self-attention with our Hilbert-Mamba block in both paths:
\begin{itemize}
    \item \textbf{Stage 1 (Main Path):} Enhanced features are fed into a Mask Decoder, composed of Hilbert-Mamba-based blocks, to generate a primary, coarse prediction.
    \item \textbf{Stage 2 (Refinement Path):} In parallel, a refinement path uses Multi-scale Context Aggregation Units (MC AU) and guidance from the Hilbert Mamba Attention \& Bank ($M_t$) to progressively refine boundaries. This fusion at each stage $j$ is defined as:
    \begin{equation}
    f_{j} = 
    \begin{cases}
        \text{Conv}(\text{Cat}(f_{j}^{t}, f_{j}^{t-1})), & j=4 \text{ (coarsest)} \\
        \text{Conv}(\text{Cat}(f_{j}^{t}, f_{j}^{t-1}, \text{UP}(f_{j+1}))), & j \in \{1, 2, 3\}
    \end{cases}
    \label{eq:decoder_fusion}
    \end{equation}
    where $f_j^t$ is the main path feature, $f_j^{t-1}$ is the memory bank feature, and $\text{UP}(\cdot)$ is a Transposed-Convolution.
    \item \textbf{Final Prediction:} The \textbf{Segmentation Head} combines both path outputs to generate the final, precise segmentation mask.
\end{itemize}

\begin{figure}[!t]

\centering

\includegraphics[width=\columnwidth]{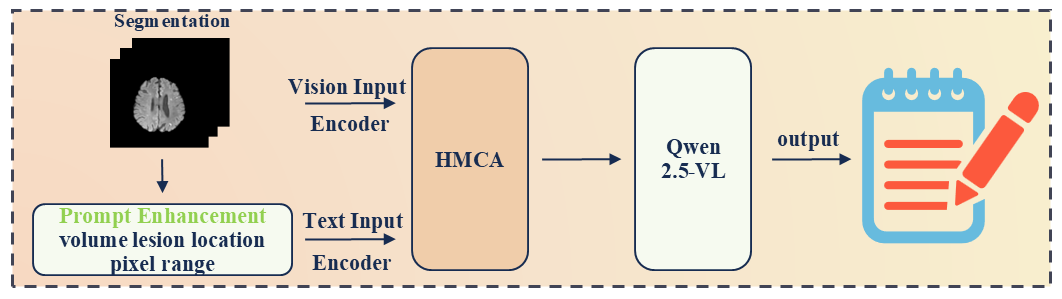}

\caption{
        An illustration of the Stage 2 Fused Prompting mechanism for VLM-based classification.
    }

\label{VLM}

\end{figure}

\subsection{Stage 2: Fused Prompting for VLM-based Classification}
In order to establish a connection between the segmentation output and the VLM's reasoning, a Fused Prompt Module has been designed (see Figure ~\ref{VLM}). The purpose of this module is to translate the visual mask and its derived textual attributes into an enhanced prompt. This provides the VLM with a comprehensive, information-dense input for accurate diagnostic reasoning.

\subsubsection{Multimodal Prompt Construction}
The module creates two token streams from the HilbertMed-SAM segmentation results:
\begin{itemize}
    \item \textbf{Visual Tokens:} A lightweight visual encoder processes the binary segmentation mask to create visual embedding tokens.
    \item \textbf{Textual Tokens:} Key attributes (e.g., lesion location, volume) are extracted from the mask and formatted as a natural language sentence. The VLM's text tokenizer then creates the textual tokens.
\end{itemize}

\subsubsection{Prompt Fusion and Training Objective}
The HMCA mechanism then merges these visual and textual token sequences into a single representation, $R_{enhanced}$. This enhanced prompt is prepended to the classification query to guide the VLM. The VLM is then fine-tuned using a composite loss function:
$$
J_{\text{VLM}} = \mathcal{L}_{\text{CE}}(Y_{\text{pred}}, Y_{\text{true}}) + \lambda \cdot \mathcal{L}_{\text{Consist}}(R_{\text{enhanced}}, Y_{\text{true}})
$$
where $\mathcal{L}_{\text{CE}}$ is the standard cross-entropy loss for classification. $\mathcal{L}_{\text{Consist}}$, a consistency loss (e.g., a contrastive loss), encourages the VLM's hidden representations from $R_{enhanced}$ to align with those from the ground-truth text. $\lambda$ is a hyperparameter balancing the two. This objective encourages the VLM to align its internal representations with the provided visual evidence, in addition to predicting the correct label.

\section{Experiments}

% ----- Table 2: SOTA Comparison for Segmentation -----
\begin{table*}[htbp]
    \centering
    \caption{Comparison of segmentation methods on brain lesion datasets. Our model, HilbertMed-SAM, achieves new SOTA performance}
    \label{tab:segmentation_sota}
    \resizebox{\textwidth}{!}{%
    \begin{tabular}{ccc cccc cccc cccc}
        \toprule
        \multirow{2}{*}{\textbf{\#}} & \multirow{2}{*}{\textbf{Method}} & \multirow{2}{*}{\textbf{Year}} & \multicolumn{4}{c}{\textbf{BraTS2021}} & \multicolumn{4}{c}{\textbf{FCD2023}} & \multicolumn{4}{c}{\textbf{Average}} \\
        %\cmidrule(lr){4-7} \cmidrule(lr){8-11} \cmidrule(lr){12-15}
        & & & Dice(\%) & IoU(\%) & Prec(\%) & Sens(\%) & Dice(\%) & IoU(\%) & Prec(\%) & Sens(\%) & Dice(\%) & IoU(\%) & Prec(\%) & Sens(\%) \\
        \midrule
        1  & UNETR      & 2022 & 52.26 & 48.09 & 56.75 & 49.32 & 26.68 & 16.09 & 33.45 & 25.56 & 37.27 & 23.76 & 39.46 & 34.32 \\
        2  & SwinUNETR    & 2021 & 53.27 & 47.14 & 54.43 & 53.33 & 28.64 & 16.30 & 29.71 & 27.86 & 37.62 & 24.06 & 33.98 & 41.91 \\
        3  & nnFormer   & 2021 & 62.36 & 54.80 & 63.50 & 61.45 & 43.88 & 30.35 & 46.91 & 45.61 & 50.12 & 35.21 & 52.83 & 46.97 \\
        4  & MixUNETR   & 2025 & 61.98 & 52.14 & 59.65 & 62.31 & 38.61 & 24.52 & 35.56 & 41.72 & 47.74 & 32.95 & 49.21 & 46.39 \\
        5  & STUNET       & 2023 & 59.80 & 51.06 & 60.32 & 57.11 & 27.42 & 16.75 & 28.81 & 26.18 & 22.54 & 15.82 & 25.13 & 28.80 \\
        6  & MedSAM       & 2024 & 68.44 & 55.08 & 69.60 & 68.21 & 56.43 & 40.77 & 51.20 & 58.61 & 66.65 & 54.27 & 64.51 & 71.46 \\
        7  & SAM Med2D  & 2024 & 71.38 & 61.46 & 73.93 & 67.35 & 57.51 & 42.96 & 57.30 & 58.91 & 67.02 & 52.16 & 65.19 & 69.34 \\
        8  & SAM 2         & 2025 & 00.03 & 00.01 & 00.02 & 00.03 & 00.03 & 00.01 & 00.01 & 00.02 & 00.01 & 00.01 & 00.02 & 00.04 \\
        9  & SAM Med3D    & 2024 & 73.56 & 64.77 & 75.89 & 71.33 & 54.01 & 37.66 & 51.09 & 55.22 & 69.34 & 56.40 & 70.77 & 67.54 \\
        10 & MedSAM 2    & 2024 & 76.17 & 68.43 & 81.33 & 75.22 & 64.22 & 50.13 & 62.78 & 67.01 & 68.57 & 57.31 & 69.65 & 66.45 \\
        11 & FS-MedSAM2  & 2024 & 78.84 & 70.71 & 79.90 & 72.11 & 65.70 & 50.98 & 61.01 & 69.53 & 65.15 & 55.07 & 65.98 & 64.75 \\
        12 & \textbf{HilbertMed-SAM(Ours)} & 2025 & \textbf{82.35} & \textbf{73.80} & \textbf{82.50} & \textbf{81.33} & \textbf{68.45} & \textbf{51.35} & \textbf{62.00} & \textbf{71.03} & \textbf{75.40} & \textbf{62.58} & \textbf{72.25} & \textbf{76.18} \\
        \bottomrule
    \end{tabular}}
\end{table*}

% ----- Table 3: VLM method comparison -----
\begin{table}[htbp]
    \centering
    \caption{Comparison of end-to-end diagnostic classification methods. Our full Hilbert-VLM framework demonstrates clear superiority.}
    \label{tab:vlm_comparison}
    \begin{tabular}{cccccc}
        \toprule
        \textbf{\#} & Method & ACC(\%) & Recall(\%) & Prec(\%) & F1 Score \\
        \midrule
        1 & R2Gen        & 54.12 & 53.56 & 57.36 & 0.55 \\
        2 & XrayGPT      & 60.22 & 59.45 & 62.75 & 0.61 \\
        3 & XrayPULSE    & 61.81 & 67.42 & 59.86 & 0.63 \\
        4 & Med-PaLM M   & 66.81 & 65.13 & 70.27 & 0.68 \\
        5 & MiniGPT4o    & 69.33 & 83.90 & 64.90 & 0.73 \\
        6 & Qwen 2.5-VL  & 67.31 & 69.79 & 80.35 & 0.75 \\
        7 & \textbf{Hilbert-VLM(Ours)} & \textbf{78.85} & \textbf{75.35} & \textbf{85.44} & \textbf{0.80} \\
        \bottomrule
    \end{tabular}
\end{table}

\subsection{Datasets and Evaluation Protocols}
We evaluated Hilbert-VLM on public datasets using consistent data splits.

\subsubsection{Segmentation Datasets}
We assessed HilbertMed-SAM on two public MRI benchmarks: \textbf{BraTS2021} for gliomas and \textbf{FCD2023} for subtle focal cortical dysplasia lesions. We used the Dice Similarity Coefficient (DSC) and 95th percentile Hausdorff Distance (HD95) as primary metrics.

\subsubsection{Diagnostic Classification Datasets}
We validated the full framework on a dataset of 4,800 annotated 2D slices (glioma, epilepsy, cerebral infarction). These were selected from the BraTS2021 and FCD2023 volumes, which directly links the segmentation and classification tasks. We measured performance using Accuracy (ACC), Precision (Prec), Recall, and F1 Score.

\subsection{Performance Evaluation}

\subsubsection{State-of-the-Art Segmentation Performance of HilbertMed-SAM}
\begin{figure}[!t]
\centering
\includegraphics[width=1.0\columnwidth]{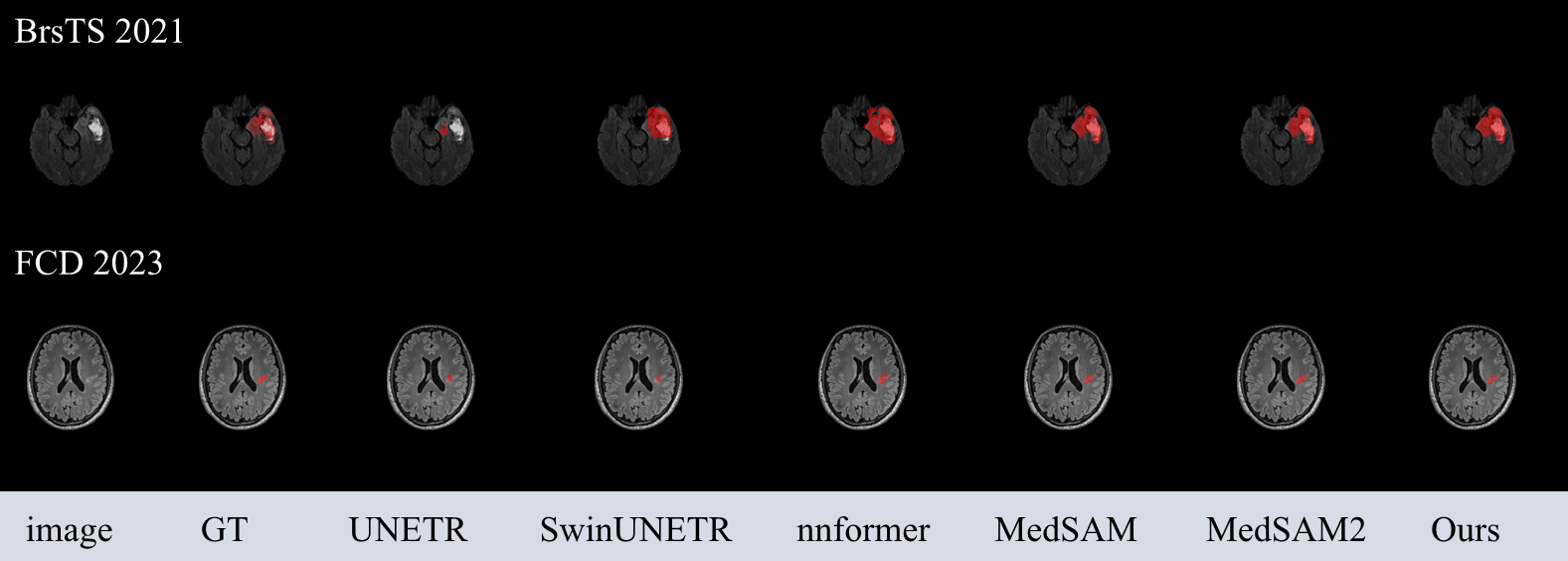}
\caption{Qualitative comparison of segmentation results from our HilbertMed-SAM against other state-of-the-art methods on representative slices from the BraTS2021 and FCD2023 datasets.}
\label{fig_slice}
\end{figure}

As presented in Table~\ref{tab:segmentation_sota}, HilbertMed-SAM establishes new SOTA results. On the BraTS2021 dataset, it achieves a top Dice score of 82.35\%, significantly outperforming prior methods. It also demonstrates excellent performance on the FCD2023 dataset, achieving the highest average Dice score of 75.40\%. The qualitative results in Figure~\ref{fig_slice} corroborate these findings, showing our model produces more complete segmentations with fewer false positives, particularly for subtle FCD lesions.

\subsubsection{End-to-End Diagnostic Performance of Hilbert-VLM}
The final diagnostic accuracy of the full Hilbert-VLM framework is detailed in Table~\ref{tab:vlm_comparison}. Our model achieved the highest ACC (78.85\%), Precision (85.44\%), and F1 Score (0.80), surpassing other leading VLM-based methods. This confirms that the synergy between HilbertMed-SAM's high-fidelity segmentation and our Fused Prompt Module translates into more reliable clinical classification.

\section{Conclusion}
This paper introduces Hilbert-VLM, a two-stage framework mitigating VLM "perceptual failures" in complex medical diagnosis. HilbertMed-SAM, re-engineered from SAM2 via a novel Hilbert-Mamba principle, handles precise lesion segmentation. A fused prompt then translates these visual findings, helping reduce VLM hallucinations and improve diagnostic reasoning. The full Hilbert-VLM framework achieves new SOTA results in segmentation and classification, showing its potential as an effective tool for clinical AI.

\bibliographystyle{IEEEtran}  % 设置参考文献格式为 IEEE 标准格式
\bibliography{references}      % 关联您的 .bib 文件 (无需写 .bib 后缀)

\end{document}